\pdfoutput=1

\documentclass[11pt]{article}

\usepackage[]{coling}

\usepackage{times}
\usepackage{latexsym}

\usepackage[T1]{fontenc}
\usepackage{fourier}
\usepackage[utf8]{inputenc}

\usepackage{microtype}

\usepackage{inconsolata}

\usepackage{graphicx}

\usepackage{times}
\usepackage{latexsym}
\usepackage{stmaryrd}
\usepackage{enumitem}
\usepackage[most]{tcolorbox}
\usepackage{bm}
\usepackage{subfigure}
\usepackage{subcaption}
\usepackage[T1]{fontenc}

\usepackage[utf8]{inputenc}
\usepackage{lipsum}
\usepackage{microtype}
\usepackage{bbm}
\usepackage{graphicx}
\usepackage{hyperref}
\usepackage{csquotes}
\usepackage{xcolor}
\usepackage{adjustbox}
\usepackage{makecell}
\usepackage{multirow}
\usepackage{amsmath}
\usepackage{amssymb}
\usepackage{inconsolata}
\usepackage{pifont} 

\usepackage{algorithm}
\usepackage[noend]{algorithmic}
\usepackage{amsmath}
\usepackage{fourier}

\usepackage{booktabs}
\usepackage{array}
\title{Unveiling Performance Challenges of Large Language Models in Low-Resource Healthcare: A Demographic Fairness Perspective}

\author{\textbf{Yue Zhou} \quad\textbf{Barbara Di Eugenio} \quad\textbf{Lu Cheng}\\University of Illinois Chicago \\\tt \{yzhou232,bdieugen,lucheng\}@uic.edu}

\begin{document}
\maketitle
\begin{abstract}
This paper studies the performance of large language models (LLMs), particularly regarding demographic fairness, in solving real-world healthcare tasks. We evaluate state-of-the-art LLMs with three prevalent learning frameworks across six diverse healthcare tasks and find significant challenges in applying LLMs to real-world healthcare tasks and persistent fairness issues across demographic groups. We also find that explicitly providing demographic information yields mixed results, while LLM's ability to infer such details raises concerns about biased health predictions. Utilizing LLMs as autonomous agents with access to up-to-date guidelines does not guarantee performance improvement. We believe these findings reveal the critical limitations of LLMs in healthcare fairness and the urgent need for specialized research in this area.~\textcolor{orange}{\bf \raisebox{0.325ex}{\resizebox{!}{1.2ex}{\warning}}WARNING: This paper contains model outputs that may be considered offensive in nature.}
\end{abstract}

\section{Introduction}

The application of Artificial Intelligence (AI)  in healthcare is almost as old as AI itself\footnote{The first issue of the journal \textit{AI in medicine} is from 1989; \url{https://www.sciencedirect.com/journal/artificial-intelligence-in-medicine/}}. Over the years, the penetration of AI techniques in healthcare has increased, from early expert systems like MYCIN \cite{shortliffe1976computer} to NLP  techniques applied to clinical notes \cite{friedman1999natural} to the current proliferation of applications of  Large Language Models (LLMs) \cite{attempt4-survey}. The assumption is that  LLMs will be equally successful in healthcare as they have been in other domains ~\cite{LLM-eval1,LLM-eval2,LLM-eval3}, especially given emerging learning frameworks such as chain-of-thought, parameter-efficient fine-tuning, and LLM as autonomous agents to address in-context reasoning, data scarcity and factual knowledge~\cite{cot,cot0,cot-paraphrase,ReAct,Reflexion, llm-agent-survey}.


Healthcare applications present unique challenges due to the complexity of knowledge involved, limited data resources, and inherent ethical considerations, including how to mitigate health disparities and achieve health equity \cite{pereira1993mean,laveist2005minority,waters2000measuring,braveman2006health,lane2017equity,ndugga2021disparities}. While recent studies have begun exploring  LLMs in medical QA, bio-medicine understanding, and disease diagnosis~\cite{attempt1-palmmed2, attempt2-biomedicine, attempt3-medicine, attempt4-survey, attempt5-raredis}, there remains a significant gap in comprehensive evaluations of LLM performance on real-world healthcare tasks, particularly as concerns  
their potential to reinforce health disparities, a crucial consideration given LLM known biases and their potential impact on patient care ~\cite{unfair1, unfair2,faireva1, faireva2}.

To address this gap and provide insights into best practices for utilizing LLMs on low-resource healthcare tasks, we present a comprehensive study examining the performance of LLMs across diverse healthcare benchmarks. Concretely, we formulate six benchmarks, including mortality, readmission, health coaching outcome prediction, and mental health diagnosis. We evaluate three state-of-the-art LLMs, GPT-4~\cite{openai}, Claude-3~\cite{claude3}, and LLaMA-3~\cite{llama3modelcard}, with prevalent frameworks: in-context learning with chain-of-thought reasoning~\cite{cot,cot0}, parameter-efficient fine-tuning~\cite{lora,qlora}, and LLM-as-agent leveraging external factual knowledge. We employ two standard fairness metrics, Demographic Parity Difference (DPD) and Equal Opportunity Difference (EOD)~\cite{fairmetric1,fairmetric2,fairmetric3} to quantify disparities across racial and gender groups, offering insights into the models' potential biases. 

Our findings show significant challenges in applying LLMs to real-world healthcare tasks. Contrary to their success in other domains, LLMs struggle to achieve high accuracy across our benchmarks, with some implementations barely surpassing random guessing. We also observe persistent fairness issues, with considerable disparities in performance across demographic groups, particularly regarding ethnicity. Notably, explicitly prompting LLMs with demographic information yields mixed results and does not consistently improve either prediction performance or fairness. We also explore LLM ability to infer demographic information from conversations and find that LLMs can deduce demographic details with serious biases, raising concerns about their potential influence on health predictions. Finally, we reveal that access to up-to-date guidelines and factual information does not guarantee accurate predictions in healthcare scenarios.

\section{Related Work}

\paragraph{LLMs in Healthcare.} This domain has recently seen a surge in the application of LLMs. Google proposed PalmMed2~\cite{attempt1-palmmed2}, an LLM in the medical domain. \citet{attempt3-medicine, attempt6-med, attempt2-biomedicine} discuss the current applications and future landscape of LLMs in medicine. \citet{attempt4-survey} offers a review of healthcare data and applications with LLMs. \citet{attempt5-raredis} explore utilizing LLMs for rare case diagnosis. \citet{attempt7-ner1, attempt7-ner2} study named entity recognition with LLMs in clinical and biomedicine settings. However, limited work exists on demographic fairness in LLMs across multiple healthcare applications. 

\paragraph{LLMs and Fairness.} 
Large language models have demonstrated considerable in-context learning abilities~\cite{cot,cot0} and parameter-efficient fine-tuning possibilities such as Low-Rank Adaptation~\cite{lora,qlora}. Recently, the use of LLMs as autonomous agents equipped with tool usage capabilities shows promising results~\cite{ReAct, Reflexion, wang2024survey}. Nonetheless, LLMs can exhibit limitations on generating unbiased and faithful output, with performance deterioration among underrepresented groups~\cite{LLM-eval2, LLM-eval1, LLM-eval3, unfair1, unfair2,faireva1, faireva2}.


\section{Datasets and Task Formulation} To facilitate evaluating LLM performance in healthcare, especially in demographic fairness, we formulate six tasks based on four healthcare datasets containing demographic information, such as age, gender, and ethnicity. We point out that public healthcare datasets with demographic information are scarce due to potential ethical and privacy concerns. 
We employ MEDQA (publicly available),  MIMIC, which is available upon request, and two others that are either partially available (a subset of the \textit{Health Coaching Dataset} is available, but not the demographic information) or not available (\textit{Bipolar Disorder and Schizophrenia Interviews}).
Our results on the publicly available ones should attest to the generalizability of our approach.


\paragraph{MIMIC-IV}\cite{mimic} is a large, publicly available healthcare dataset containing hospital health records. Taking the patients' clinical notes as input, we formulate two tasks: \textbf{one-year mortality} prediction and \textbf{90-day readmission} prediction, which emulate patient outcome prediction in real-world settings. The input note and label pairs are created by joining three tables in the MIMIC database: `\textit{patients},' `\textit{admissions},' and `\textit{discharge}.' The note mainly contains sections including chief complaint, history of present illness, past medical history, and lab results. We removed the discharge instruction from the note in the input to our models since they can reveal direct information on mortality/readmission (e.g., Hospice).

\paragraph{Health Coaching Datasets} Dialogues are inherently unique in fairness evaluation since they often contain implicit demographic cues. This raises questions about how LLMs handle these subtleties and whether their responses might exhibit unfairness. The health coaching datasets~\cite{HC-1,HC-3} comprise SMS conversations between patients and certified health coaches over several weeks, focusing on creating and accomplishing S.M.A.R.T. goals to promote health behavior changes \cite{doran1981there}. Each week, the conversation starts with a \textit{goal setting stage}, where the coach and the patient discuss and create a concrete and measurable goal for physical activities. Then, the coach follows up on the patient's progress and maintains engagement, which is called the \textit{goal implementation stage}. We formulate another patient \textbf{outcome prediction} task, which predicts whether the patient will accomplish the next goal based on the dialogue history over the past two weeks and the current goal-setting stage.

\paragraph{Bipolar Disorder and Schizophrenia Interviews}\cite{SPSS} contains transcribed interactions between a trained clinician and outpatients with schizophrenia, bipolar disorder, and healthy controls. It includes two scenes: (1) \textbf{Meeting New Neighbor}: the participant is asked to imagine an affiliative scene and converse with the interviewee (role-playing as the new neighbor) as they have just moved into the neighborhood. (2) \textbf{Complaining to a Landlord}: A confrontational scene where the participant role-plays the tenant and complains to the landlord (role-played by the interviewee) about issues such as pipe leakage. These contrasting scenarios aim to assess mental status in both friendly and stressful situations. We utilize both scenes and ask the model to predict the outpatients' cohort based on the conversations. We aim to test the LLM diagnosis ability in real-world settings and whether they can potentially identify the linguistic cues in the conversations that specify mental illness. This task, again, aims to investigate how LLMs process and respond to implicit demographic cues within dialogues and introduce potential unfairness in their outputs.

\paragraph{Medical Question Answering Dataset}\cite{medqa} contains multiple-choice medical problems collected from medical board exams. We use the English subsection adapted from the USMLE (United States Medical Licensing Examination) and only select the test problems targeting specific ethnic groups. We use this dataset to study the performance inequality in LLMs in typical-case diagnoses for specific demographic subgroups. While commonly used to evaluate LLM performance in healthcare, this dataset can understate the challenges of real-world medical data. We hope it can serve as a comparison, highlighting the gap between controlled benchmarks and the complexity of real-world healthcare applications.

\begin{figure*}[!ht]
    \centering 
    \includegraphics[width=\textwidth]{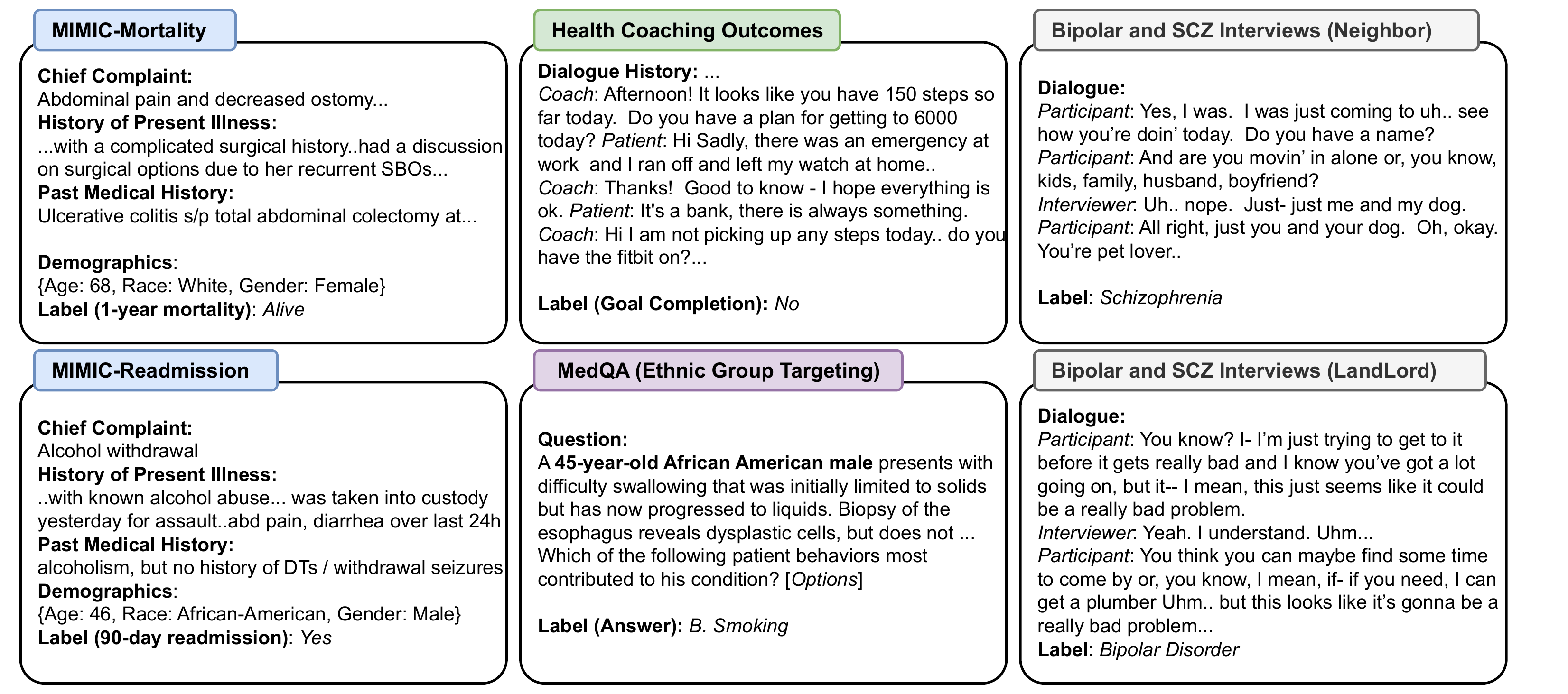}
    \caption{Overview of the six health benchmarks, with illustrated examples.}
    \label{fig:the6} 
\end{figure*}

Figure~\ref{fig:the6} provides data samples from the six healthcare benchmarks. Statistics on data split are shown in Table~\ref{tab:stats}. The small size of the Health Coaching and Schizophrenia/Bipolar datasets reflect the inherent scarcity of authentic data prevalent in patient-facing healthcare applications. For tasks with access to larger databases (MIMIC and MedQA), we intentionally capped the training data at 5000 examples to mimic similar low-resource conditions. We sample each dataset such that the classes $C$, the demographic attributes $Z$, and the distribution $P(C = c | Z = z)$ are roughly balanced.

\begin{table}[!t]
\centering
\begin{adjustbox}{width=0.8\columnwidth}

\begin{tabular}{lll}\hline\hline 
\textbf{Task }         & \textbf{SFT} & \textbf{Test} \\\hline
MIMIC-Mortality           & 5000 & 500 \\
MIMIC-Readmission         & 5000 & 500 \\
Health Coaching           & 120  & 60  \\
MedQA                     & 5000 & 175 \\
SCZ and Bipolar: Neighbor & 190  & 261 \\
SCZ and Bipolar: Landlord & 188  & 261 \\\hline
\end{tabular}

\end{adjustbox}
\caption{Data splits for fine-tuning (SFT) and testing of the tasks.} 
\label{tab:stats}
\end{table}

 We find it challenging to gather publicly available healthcare corpora where demographic information is also available, due to ethical concerns and privacy issues in the healthcare domain.  We posit that there is a need for more open-source, well-formatted healthcare data to facilitate research on AI fairness.

\section{Baselines} Our study aims to evaluate the performance of trendy frameworks utilizing large language models in real-world, low-resource healthcare settings. We seek to provide insights into best practices for leveraging LLMs when building applications under these constraints. To this end, we evaluate the performance of LLMs using three representative frameworks:

\noindent \textbf{$\bullet$ In-Context Learning (ICL) with Chain-of-Thought}~enhances LLM inherent reasoning capabilities by prompting the LLM to provide a step-by-step reasoning chain~\cite{cot,cot0, cot-paraphrase}. We implement two schemes: \textbf{(1)} \textbf{Zero-Shot Chain-of-Thought (CoT)}, which appends ``\emph{Let's think step by step.}'' to the question text; and \textbf{(2)} \textbf{N-Shot CoT}, where we append four to eight-shot in-context examples with CoT to the LLM when solving the problems. The examples are demographically balanced following~\cite{fairmetric2}. We report the best performance between Zero-Shot and N-Shot in this setting. Since previous work found weak evidence on prompting demographic information to improve fairness~\cite{fairmetric2}, we provide baselines with and without explicit demographic information in the input in this framework.

\noindent \textbf{$\bullet$ Parameter Efficient Fine-Tuning (PEFT)} is a set of techniques that adapt pre-trained language models to downstream tasks by updating only a small portion of the parameters, reducing computational costs and storage requirements while maintaining performance. Given the resource constraints, we specifically fine-tune LLaMA-3 with \textbf{Low-Rank Adaptation (LoRA)}~\cite{lora}, where the pre-trained model weights are untouched, yet small-scale trainable rank decomposition matrices are injected. 

\begin{figure}[!t]
    \centering 
    \includegraphics[width=\columnwidth]{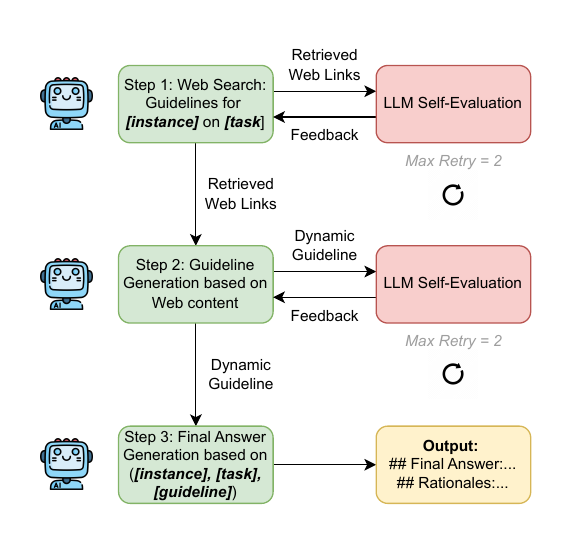}
    \caption{An overview of the LLM as Agent framework, including search and generating guidelines for the underlying data instance and task, and generating the final answer referencing the guidelines.}
    \label{fig:agent_overview}
\end{figure}

\noindent \textbf{$\bullet$ LLM as Agents} enhances LLMs with specialized modules for planning and tool usage, enabling them to solve complex tasks beyond pre-trained knowledge. In the paper, we propose a simplified pipeline based on \textbf{ReAct}~\cite{ReAct} and \textbf{Reflexion}~\cite{Reflexion}. Concretely, for each task, we prompt the LLM to web search for the latest guidelines for analyzing the underlying \texttt{[example]} on the \texttt{[task]}. Then, we prompt the LLM to generate a concise guide based on the retrieved top 10 most relevant Google search results. Finally, the LLM generates predictions based on the question, the example, and the generated guide. We additionally allow a maximum of two re-attempts for the first two steps based on LLM's self-evaluation. Figure~\ref{fig:agent_overview} shows an overview of the framework.

\section{Metrics for Fairness} Existing literature predominately adopts two metrics to evaluate the demographic fairness of the model prediction~\cite{fairmetric1,fairmetric2,fairmetric3}. The first metric is called \textit{Statistical Parity} or \textit{Demographic Parity}. Statistical parity is achieved when favorable decision outcomes are unrelated to the protected attributes. The rationale is to test whether the model treats various subgroups similarly. Take fraud detection as an example; the model should output ``good credit'' with a similar chance for both males and females. Note that it does not consider the ground truth label. Consider the sensitive/demographic attribute $Z$ and the predicted outcome $\hat{Y}$, the (one-vs-all) \textbf{Demographic Parity Difference (DPD)} for subgroup $z_i$ can be defined as:

$$PDP = P(\hat{Y}=1|Z = z_i) - P(\hat{Y}=1|Z \neq z_i)$$

\noindent This metric may pose challenges when assessing model performance in healthcare applications, as the attribute $Z$ could be a prior factor influencing model predictions. For instance, when predicting a patient's one-year mortality, age may significantly influence risk, with individuals above the age of 90 facing greater risk compared to those below. Consequently, an LLM which obtained such knowledge during pre-training may be more likely to predict mortality for patients above 90 years old. Nonetheless, we include this metric as it provides valuable insights into the model's prediction tendencies across different demographic groups in healthcare contexts and is crucial for understanding potential biases.

The second metric, \textit{Equality of Opportunity}, evaluates model fairness based on the ground truth labels. It indicates that different subgroups should have an equal likelihood of being accurately classified by the model. One way to formulate the metric is to measure the true positive rates of class $Y$ across various subgroups. We report the \textbf{Equal Opportunity Difference (EOD)} as:

\begin{multline*}
EOD = P(\hat{Y} = 1|Y = 1, Z = z_i) \\ - P(\hat{Y} = 1|Y = 1, Z \neq z_i)
\end{multline*}


\noindent Note the definition of favorable attributes in healthcare is more nuanced than in other domains like fraud detection or tweet classification. While ``good credit'' or ``non-toxic'' are straightforward favorable attributes in those fields, healthcare scenarios often have context-dependent favorable classes. However, for clarity and consistency in our main experiments, we define favorable attributes by \textit{any positive health indicators} across different tasks. These include, for example, Low Mortality Risk out of \{Low Mortality Risk, High Mortality Risk\} in mortality prediction and Healthy Control group out of \{SCZ, Bipolar, Healthy Control\} in SCZ and Bipolar Interviews. We report global accuracy and accuracy per demographic group instead of PDP and EOD for the MedQA task, which involves open-ended question answering. 

\section{Experiments}

In this section, we describe our experiment results evaluating the effectiveness of LLMs in solving real-world healthcare tasks with various frameworks and settings, as well as additional discussions on demographic awareness and qualitative examples.

\subsection{Experimental Settings}

\paragraph{Language Models} We utilize three state-of-the-art large language models for evaluation, including two closed-source models, OpenAI GPT-4~\cite{openai} and Claude-3 (Sonnet)~\cite{claude3}, and one open-source model, LLaMA-3 (8b)~\cite{llama3modelcard}. In compliance with the responsible use guidelines for MIMIC data with online services, we utilize the Azure OpenAI service for GPT-4 and opt out of human data review.\footnote{\href{https://physionet.org/news/post/gpt-responsible-use}{Responsible use of MIMIC data}} We have also ensured that our usage of Claude-3 adheres to the agreement.\footnote{\href{https://support.anthropic.com/en/articles/7996885-how-do-you-use-personal-data-in-model-training}{Anthropic's data usage policy}} The LLaMA-3 model is run locally on our machines. The Schizophrenia and Bipolar dataset is the only dataset in our study that is not publicly available and requires approval from an Institutional Review Board (IRB). 


\paragraph{Implementation Details} 

For fine-tuning, we employed LoRA with a rank of $8$ across all trainable layers. We use a dropout rate of $0.1$, a learning rate $1e^{-5}$, and a batch size of $8$ for all experiments. Our implementation adheres to the recommendations outlined in QLoRA \cite{qlora}, except for the LoRA scaling factor (Alpha), which is set equal to the LoRA rank. We choose the temperature $T = 0.3$ for all three language models for inference. The full implementation details and prompt templates used in the experiments are available in Appendix~\ref{sec:app1}.

\subsection{Main Results}

Table~\ref{tab:res1} shows the accuracy results for six healthcare tasks using different LLM frameworks. The numbers outside parentheses represent accuracy without explicit demographic information as input, while those inside parentheses show results when demographic information is explicitly prompted to the LLMs. There are several key observations: \ding{182} Despite their impressive performance in various domains, LLMs struggle with real-world healthcare tasks across all prevalent frameworks. Many implementations in Readmission, Neighbor, and Landlord barely surpass random guess baselines. The claims that LLMs can easily solve classification tasks with few examples are unsupported and inconsistent with our findings regarding real-world healthcare applications. \ding{183} While closed-source, large-scale LLMs generally outperform open-source, smaller models in in-context learning, the most effective framework varies by task. For instance, the "Schizophrenia and bipolar" diagnosis tasks achieve the best results with fine-tuning despite many fewer training examples compared to MIMIC-based tasks. In contrast, for the MIMIC and health coaching tasks,   in-context learning achieves the best performance. \ding{184} The LLM-as-Agent approach shows mixed results across tasks. It excels in MedQA, presumably due to its ability to search online for open-book guidelines for USMLE questions. However, it underperforms in real-world healthcare applications despite generating seemingly convincing thought processes. The following subsection will present qualitative examples to illustrate these findings. \ding{185} Explicitly prompting LLMs with demographic information does not necessarily improve performance. The impact varies depending on both the specific task and the LLM used. We present further demographic fairness results next.

Table~\ref{tab:res2} shows Demographic Parity Difference (DPD) and Equal Opportunity Difference (EOD) across six tasks, with results inside parentheses indicating explicit demographic prompts. For brevity, we mainly focus on White vs. African American and Female vs. Male comparisons. The PDP/EOD metrics are calculated as (African American - White) for race and (Female - Male) for gender, and the +/- indicates the sign of the difference.
To interpret the results, consider the Demographic Parity Difference (DPD) and Equal Opportunity Difference (EOD) metrics. A DPD value of -8.2 for African Americans in the mortality task indicates that the model is 8.2\% less likely to predict a favorable outcome (e.g., low mortality risk) for this group compared to White patients, regardless of the ground truth. Similarly, an EOD value of -3.5 for African Americans signifies that the model's true positive rate in predicting favorable outcomes is 3.5\% lower for this group, highlighting a performance disparity. There are several key observations: \ding{182} Unfairness exists across all tasks, frameworks, and demographics, with racial disparities more prominent than gender disparities. \ding{183} LLMs consistently predict less favorable outcomes for African American patients, while a lower Equality of Opportunity for African Americans is observed in most tasks, except health coaching. \ding{184} Explicitly prompting demographic information yields mixed results on fairness. DPD mostly improves for GPT-4 but not for other models. EOD is less influenced by demographic prompts compared to DPD. \ding{185} Fine-tuning's impact on fairness varies by task, improving for some (readmission, neighbor, landlord) while worsening for others (health coaching, mortality). The agent approach can mitigate unfairness in certain cases. Note that LLaMA's all-zero results for landlord and neighbor tasks stem from blindly predicting all participants as having schizophrenia. Finally, all LLMs demonstrate a discrepancy in in-context learning performance on MedQA regarding racial group-targeting questions, see Table~\ref{tab:res3}. 

We further use the mortality dataset as an example to showcase fairness results across diverse demographic subgroups in gender, age, and race in Appendix~\ref{sec:app2}. We can observe that LLMs predict high mortality risks for the geriatric age group and African Americans and lower prediction performance for these groups.


\begin{table*}[ht!]
\centering
\begin{adjustbox}{width=1\textwidth}

\begin{tabular}{|l|l|l|l|l|l|l|l|}\hline\hline
Baselines                     & Backbones     & Mortality   & Readmission & MedQA    & HealthCo          & Neighbor & Landlord \\ \hline
Random Guess        & -         & 50.0 & 50.0 & 20.0 & 50.0 & 33.3             & 33.3            \\
ICL/Chain-of-Thought          & GPT-4         & 77.0 (\textbf{79.0}) & 55.3 (56.8) & - (68.6) & 76.7 (76.7) & 41.0 (39.1)             & 38.3 (36.8)             \\
ICL/Chain-of-Thought          & Claude-3      & 72.6 (50.8) & 52.9 (\textbf{57.6}) & - (65.7) & \textbf{80.0} (\textbf{80.0}) & 37.5 (38.3)             & 37.2 (34.1)             \\
ICL/Chain-of-Thought          & LLaMA-3       & 73.0 (72.2) & 53.3 (55.3) & - (59.4) & 70.0 (70.0) & 33.0 (33.3)   & 34.1 (33.7)   \\
Supervised Fine-Tuning & LLaMA-3       & 68.1        & 48.2        & 71.4     & 70.0          & \textbf{49.4}          & \textbf{41.4}          \\
LLM as Agent                  & GPT-4 & 66.7        & 53.3        & \textbf{86.9}     & 70.0          & 36.8          & 36.8  \\ \hline       
\end{tabular}

\end{adjustbox}
\caption{Global accuracy across the six tasks with various LLM frameworks. ICL Results with and without explicit demographic prompts are inside and outside parentheses, respectively.} 
\label{tab:res1}
\end{table*}

\begin{table*}[ht!]
\centering
\begin{adjustbox}{width=1\textwidth}
\begin{tabular}{|l|l|l|l|l|l|}\hline\hline
\textbf{Task}           & \textbf{Setting}       & \multicolumn{2}{c}{\textbf{DPD}}       & \multicolumn{2}{c}{\textbf{EOD}}       \\ \hline
&               & \textbf{African-American - White} & \textbf{Female - Male} & \textbf{African-American - White} & \textbf{Female - Male} \\ \hline
& COT-GPT-4     & -8.2 (-7.4)   & -5.1 (-2.7)   & -3.5 (-4.4)   & -4.6 (-5.3)   \\
& COT-Claude-3  & -11.5 (-6.4)  & -8.7 (+1.5)   & -3.8 (-3.8)   & -4.5 (-7.2)   \\
\textbf{Mortality}      & COT-LLaMA-3   & -8.3 (-17.6)  & -2.8 (-3.0)   & -4.2 (-8.0)   & -5.9 (-5.2)   \\
& SFT (LoRA)    & -14.3         & -12.7         & -9.4          & -5.3          \\
& LLM as Agent & +2.1          & -2.1          & +2.4          & +2.8          \\ \hline
& COT-GPT-4     & -10.5 (-7.1)  & -0.5 (-0.6)   & -9.8 (-7.1)   & -1.9 (-1.1)   \\
& COT-Claude-3  & -7.9 (-12.2)  & -2.8 (+6.4)   & -3.1 (-2.6)   & -6.1 (-1.6)   \\
\textbf{Readmission }   & COT-LLaMA-3   & -9.3 (-26.7)  & -1.3 (-0.9)   & -0.5 (-7.3)   & -2.3 (-3.1)   \\
& SFT (LoRA)    & +0.5          & +0.6          & +4.9          & +1.1          \\
& LLM as Agent & -12.1         & +4.2          & -6.8          & +4.6          \\ \hline

& COT-GPT-4     & -7.5 (-5.7)   & +3.8 (+3.6)   & -1.5 (-1.5)   & -9.0 (-9.0)   \\
& COT-Claude-3  & -7.9 (-7.3)   & +4.1 (-3.2)   & -0.1 (-0.9)   & -7.2 (-9.7)   \\
\textbf{Neighbor }      & COT-LLaMA-3   & 0.0 (0.0)*    & 0.0 (0.0)*    & 0.0 (0.0)*    & 0.0 (0.0)*    \\
& SFT (LoRA)    & -4.8          & +7.1          & -4.5          & +6.5          \\
& Agent (LLaMA) & -6.7          & -3.6          & -2.5          & -8.6          \\ \hline
& COT-GPT-4     & -10.3 (-4.7)  & +8.5 (+6.1)   & -0.7 (0.0)    & -9.9 (-8.9)   \\
& COT-Claude-3  & -4.1 (-20.2)  & +0.8 (+0.9)   & +2.4 (-4.9)   & -11.4 (-10.4) \\
\textbf{Landlord }      & COT-LLaMA-3   & 0.0 (0.0)*    & 0.0 (0.0)*    & 0.0 (0.0)*    & 0.0 (0.0)*    \\
& SFT (LoRA)    & -0.8          & -0.1          & +2.5          & -4.7          \\
& Agent (LLaMA) & -5.2          & +3.7          & +2.2          & -8.7  \\ \hline
&               & \textbf{African-American - Hispanic} & \textbf{Female - Male} & \textbf{African-American - Hispanic} & \textbf{Female - Male} \\ \hline
& COT-GPT-4     & -4.2 (-4.2)   & -4.3 (-4.3)   & +15.0 (+15.0) & -16.1 (-16.1) \\
& COT-Claude-3  & -8.3 (-8.3)   & -8.7 (-8.7)   & +15.0 (+15.0) & -16.1 (-16.1) \\
\textbf{HealthCoaching} & COT-LLaMA-3   & -12.5 (-12.5) & +5.6 (+5.6)   & -6.7 (-6.7)   & -6.2 (-6.2)   \\
& SFT (LoRA)    & -13.3         & -1.8          & +18.3         & -19.2         \\
& LLM as Agent & -4.2          & +14.3         & +10.8         & -1.8          \\ \hline

\hline
\end{tabular}
\end{adjustbox}
\caption{Demographic Parity Difference (PDP) and Equal Opportunity Difference (EOD) across the six tasks. Results with and without explicit demographic prompts are inside and outside parentheses, respectively. The PDP/EOD metrics are calculated as (African American - White) for race and (Female - Male) for gender, and the +/- indicates the sign of the difference.}
\label{tab:res2}
\end{table*}

\begin{table}[]
\centering
\begin{adjustbox}{width=1\columnwidth}

\begin{tabular}{lcccc} \hline\hline
\textbf{Baselines}     & \multicolumn{4}{c}{\textbf{Accuracy}}     \\ \hline
              & \textbf{African-American} & \textbf{Asian} & \textbf{Hispanic} & \textbf{White} \\ \cline{2-5}
COT-GPT-4     & 62    & 60    & 68       & 84    \\
COT-Claude-3  & 64    & 62    & 60       & 74    \\
COT-LLaMA-3   & 60    & 56    & 52       & 66    \\
SFT (LoRA)    & 86    & 62    & 52       & 76    \\
Agent (GPT-4) & 86    & 86    & 88       & 88   \\ \hline
\end{tabular}

\end{adjustbox}
\caption{Accuracy across different race-targeting MedQA problems.}
\label{tab:res3}
\end{table}

\subsection{Additional Studies}

\paragraph{Demographic Awareness and Biases} An intriguing question arises regarding the risk that LLMs  infer irrelevant or wrong demographic information from conversational data and hence,  that they introduce bias into health outcome predictions. To explore this, we ask three LLMs to identify the patient's race, given the conversation. Table~\ref{tab:res4} shows that GPT-4 achieved the highest accuracy in predicting race information, followed by LLaMA-3, outperforming random guessing. Notably, Claude-3 refused to make race predictions, stating, \textit{``I would prefer not to speculate about the participant's race or ethnicity.''} GPT-4 exhibited similar reluctance for the health coaching dataset. However, refusal to answer does not necessarily indicate an inability to do so or a lack of potential biases. Table~\ref{tab:qa_big} provides a qualitative example of LLaMA's predictions with rationales, which heavily contain stereotypical and biased assumptions (even those based on AAV (African American Vernacular)  characteristics are incorrect, according to a sociolinguistics expert on AAV we consulted\footnote{Detailed consultation refer to Appendix C.}). While the model appears to predict schizophrenia (SCZ) based on unbiased criteria independent of race, it is unclear whether racial biases subtly influence these diagnoses. In fact, we observed a high Demographic Parity Difference in schizophrenia predictions for African-American individuals. This raises serious concerns about the underlying biases that LLMs may possess and how these biases could influence healthcare-related predictions and decision-making processes.

\begin{table*}[!ht]
\begin{adjustbox}{width=\textwidth}
\begin{tabular}{lllll}
\hline\hline
\textbf{Correct}  & \textbf{Output}   & \textbf{Task}    & \textbf{Model}  \\ \hline
Healthy & \makecell[cl]{1. Linguistic features: The participant's language use, such as the use of "ain't" and "gonna" \\  is more commonly associated with African American Vernacular English (AAVE).\\ 2. Tone and demeanor: The participant's tone and demeanor throughout the conversation, \\ such as \textbf{expressing frustration and anger} when discussing the leak in their apartment, are \\ consistent with the cultural norms of African American individuals.\\
3. Language patterns: ..such as using short sentences and a more informal tone, are also AAVE.\\
4. Lack of formal language use: ..does not use formal language or attempt to articulate \\ their concerns in a more sophisticated manner, is consistent with the cultural norms of \\ some African American individuals.
}                            & \makecell[cl]{Race\\  Inference}          & \makecell[cl]{LLaMA-3}         \\\hline

Healthy & \makecell[cl]{\textbf{Answer}: Schizophrenia\\
\textbf{Rationale}: 1. Disorganized and illogical speech patterns...2. Lack of coherence and clarity...\\
3. Irritability and agitation...
4. Disregard for social norms and conventions...
}                            & \makecell[cl]{Diagnosis}          & \makecell[cl]{LLaMA-3}         \\\hline

C. 46XX & \makecell[cl]{\textbf{Answer:} C. 46XX \\
\textbf{Rationale}: The clinical presentation and ultrasound findings are indicative of a molar pregnancy. \\ The markedly elevated HCG levels and the "snowstorm" appearance on ultrasound strongly \\ suggest a complete molar pregnancy. Complete molar pregnancies most commonly have \\ a 46XX karyotype, resulting from the fertilization of an empty ovum by ...\\ \textbf{Retrieved Links}: \url{https://www.ncbi.nlm.nih.gov/books/NBK459155/} \\
\url{https://step2.medbullets.com/gynecology/121751/hydatidiform-mole}
}                          & \makecell[cl]{QA}          & \makecell[cl]{LLM \\ as \\ Agent}         \\\hline

\makecell[cl]{Bipolar\\Disorder} & \makecell[cl]{\textit{Reasoning through retrieved guidelines for SCZ (Bold Text)} \\ - \textbf{Disorganized Speech}: The participant repeats phrases like ``it's getting worse'' \\
- \textbf{Reduced Complexity}: straightforward and repetitive sentence structures, \\ such as ``I'm- I'm- I'm gonna take’''\\
- \textbf{Limited Vocabulary}: repeating the same words like ``worse,’’ ``leaking,'' and ``important.''\\
- \textbf{Poverty of Speech}: The participant's responses are often brief and lack depth, such as ``Huh? ''\\ \textbf{Answer}: Schizophrenia\\}                         & \makecell[cl]{Diagnosis}          & \makecell[cl]{LLM \\ as \\ Agent}         \\\hline

\end{tabular}
\end{adjustbox}
\caption{Qualitative examples of model outputs on health outcome prediction and race inference. }
\label{tab:qa_big}
\end{table*}

\begin{table}[]
\centering
\begin{adjustbox}{width=0.97\columnwidth}

\begin{tabular}{llll} \hline\hline
        & \textbf{HealthCoaching} & \textbf{Neighbor} & \textbf{Landlord} \\ \hline
        
Random    & 33.3                 & 50.0             & 50.0                 \\
GPT-4    & Refusal                 & 75.2              & 78.8                 \\
Claude3 & Refusal                 & Refusal           & Refusal                 \\
LLaMA-3  & 40.0                      & 59.6              & 61.7   \\ \hline             
\end{tabular}

\end{adjustbox}
\caption{Accuracy of predicting patient's race from conversations.}
\label{tab:res4}
\end{table}

\paragraph{Agent and Factual Knowledge Retrieval in Healthcare}
One potential advantage of using LLMs as agents equipped with tool usage capabilities is to retrieve external facts and knowledge to guide predictions rather than relying solely on potentially hallucination-prone pre-trained knowledge. This approach yields impressive performance in solving MedQA questions. We hypothesize that the performance stems from the LLM web search for direct guidelines, particularly for questions that require memorization rather than complex reasoning. Table~\ref{tab:qa_big} provides an example where the LLM agent directly located guideline links. 

However, our findings suggest that access to up-to-date guidelines and factual information does not necessarily guarantee accurate final predictions. Table~\ref{tab:qa_big} illustrates a factual guideline retrieved from the latest research yet incorrect reasoning from linguistic cues. The LLM erroneously overemphasized fragments and other speech patterns and thus predicted the patient as having schizophrenia, failing to account for the fact that these were spoken dialogue transcripts (despite this being explicitly stated in the prompt). This example highlights the challenges in applying retrieved knowledge appropriately and the potential for misinterpretation even when given access to current and factual information in healthcare.

\section{Conclusions and Future Work}

We explored LLM performance and demographic fairness across diverse healthcare tasks. Our experimental results highlight LLM difficulties in solving real-world healthcare tasks and significant disparities across demographic groups. We showed that explicitly providing demographic information to LLMs does not guarantee improved performance or fairness. While LLM-as-agent can retrieve factual knowledge, the reasoning processes may still lead to inaccurate conclusions. We also observed potential biases in race inference, which could influence health outcome predictions. These findings highlight the urgent need for future research addressing LLM fairness and reliability in healthcare.

\section*{Limitations}

While our study provides valuable insights into the performance disparities of LLMs across demographic groups in healthcare tasks, we still need to identify a systematic strategy to mitigate biases and improve fairness across different healthcare scenarios. Additionally, investigating the relationship between LLM demographic awareness and the fairness of health outcome predictions could yield noteworthy insights, particularly given our observations of biases in demographic inference and demographic parity differences.

\section*{Ethics Statement}
This study examined LLM performance across diverse healthcare tasks, revealing disparities in accuracy and fairness among demographic groups. Our findings highlight the potential risks of propagating LLM unfairness and the need for rigorous fairness assessments and bias mitigation in healthcare. Future work should focus on developing LLMs for equitable outcomes in healthcare.

\section*{Acknowledgement} The work is supported by the National Science Foundation under Grants IIS-1838770 and IIS-2312862.

\bibliography{custom}
\newpage
\appendix

\section{Prompt Templates}
\label{sec:app1}

We list the prompt templates used in the paper below. 
\begin{tcolorbox}[
    colback=white,     
    colframe=gray!75,
    title=Question Formulation Prompt
]
\textbf{Mortality / Readmission}: Assess the 1-year mortality risk (low or high) / 90-day readmission chance for the patient based on the given de-identified clinical note. Using medical evidence only. If uncertain, provide your best evidence-based estimate. \texttt{[Note]}

\vspace{0.5em}

\textbf{Health Coaching}: Given the following conversation between the patient and the health coach, predict the patient's outcome regarding goal achievement in the following week.
Will the patient be most likely to complete the goal in the following week based on the conversation history?

\texttt{[Conversation]}

\vspace{0.5em}

\textbf{MedQA}: Answer the following USMLE question with medical evidence only. No assumptions. If unsure, give your best evidence-based guess. \texttt{[question]} \texttt{[options]}

\vspace{0.5em}

\textbf{Neighbor Scene}: The following is a transcribed conversation from an audio recording between a participant and an interviewer, who is a trained psychologist. \texttt{[scene]}: \textit{In this scenario, the participant imagines they have just moved into a new neighborhood and must introduce themselves to a new neighbor. }The purpose of this exercise is to gather information about the participant's mental status through their linguistic cues. The participant belongs to one of three groups: individuals with schizophrenia, those with bipolar disorder, or healthy controls. Based on the dialogue, which group is the participant most likely to belong to?

\texttt{[Conversation]}

\vspace{0.5em}

\textbf{Landlord Scene}: Same as above except for \texttt{[scene]}: \textit{In this confrontational scenario, the participant imagines having a leaky pipe in their apartment that has not been fixed for a while,
and they need to complain to their landlord and get it fixed.}
\end{tcolorbox}

\begin{tcolorbox}[
    colback=white,     
    colframe=gray!75,
    title=Output Formatting and Other
]

\textbf{Output format}:

\#\#\# Answer: \texttt{[Task-Dependent Choices]}

Then, provide your rationale.

\#\#\# Rationales: ...

\vspace{1em}

Each task prompt contains optional demographic information between the instruction and the data example:
\texttt{[Optional Demographic Information]}
\end{tcolorbox}

\begin{tcolorbox}[
    colback=white,     
    colframe=gray!75,
    title=Agent Prompt
]

To facilitate implementation, we utilized the third-party package Crewai~\url{https://github.com/crewAIInc/crewAI} for autonomous agent prompting.

\textbf{Web Search}:
\textit{Description}: Search for the latest research, guidelines, or expert recommendations on analyzing \texttt{[task]} based on \texttt{[example]}. \textit{Expected Output}: Provide a concise summary (within 200 words) of key points to aid in analysis. List each point with its rationale in bullet form. \texttt{[Tools = Web Search]}

\vspace{0.5em}

\textbf{Analysis}:
Using the provided guidelines, analyze \texttt{[example]} and predict \texttt{[task]} (varies by \texttt{[Question Formulation Prompt]} and \texttt{[Output Formatting]})

\end{tcolorbox}


\section{Additional Results}
\label{sec:app2}

\begin{figure*}[!t]
    \centering 
    \includegraphics[width=0.92\textwidth]{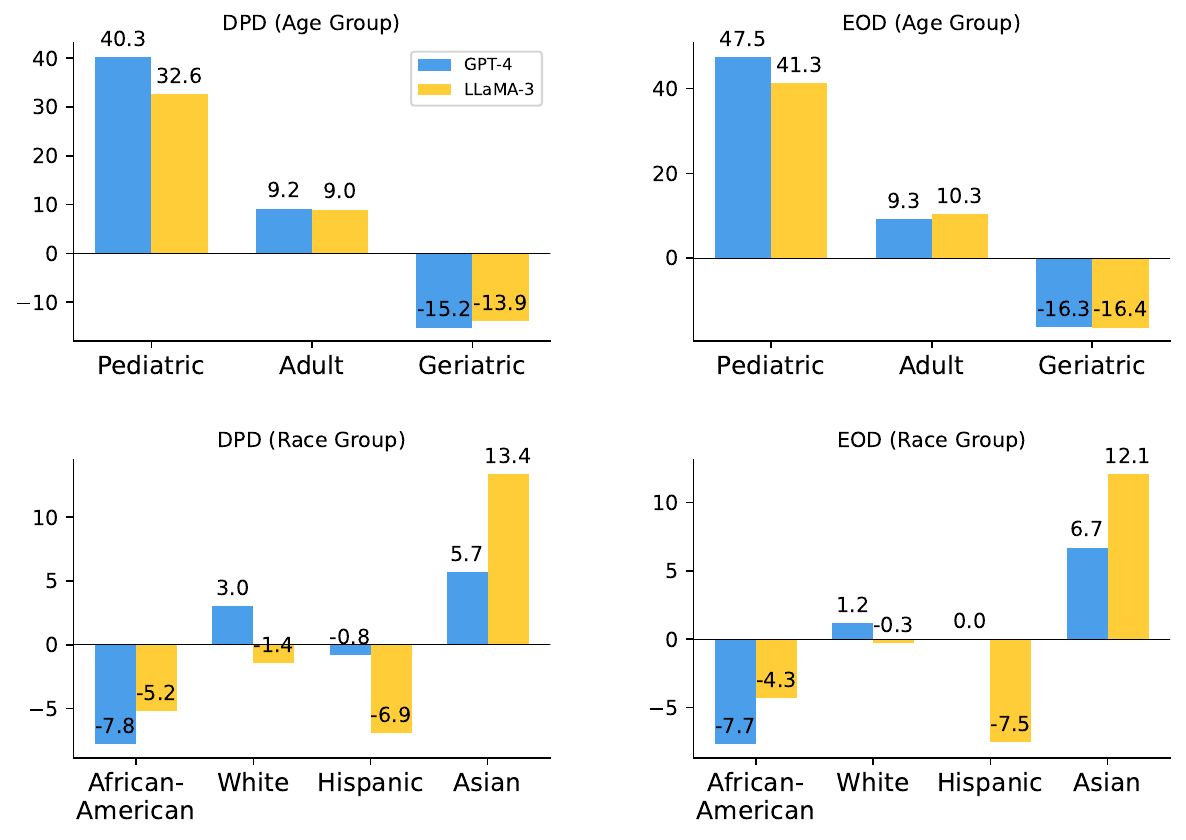}
    \caption{PDP and EOD results for more demographic subgroups (one-vs-all) for the mortality prediction task.}
    \label{fig:more_compare}
\end{figure*}

We show more fairness discrepancy across demographic subgroups in mortality prediction in Figure~\ref{fig:more_compare}. The Demographic Parity Difference (DPD) and Equal Opportunity Difference (EOD) are calculated using a one-vs-all approach. Both GPT-4 and LLaMA-3 exhibit similar bias patterns: they are more likely to predict high mortality risks for the geriatric age group and African Americans. Additionally, these models demonstrate lower prediction performance (True Positive Rate) for these groups. These findings highlight the persistent challenges in achieving LLM fairness across different demographic subgroups in healthcare settings.

\section{Sociolinguistic Consultation}

We consulted with a sociolinguist regarding language model outputs that attempt to infer demographic characteristics from conversational patterns for diagnosis. The consultation revealed significant concerns about linguistic stereotyping in current LLMs.

\paragraph{Misattribution of Common Linguistic Features} Features like ``ain't'' and ``gonna,'' which LLMs often flag as African American Vernacular English (AAVE), are prevalent across multiple dialects. The expert notes that while ``ain't'' showed some demographic correlation in specific contexts (e.g., Oak Park school study, Chicago area), it is not unique to AAVE. Similarly, ``gonna'' is a common informal contraction across all English dialects.

\paragraph{Problematic Behavioral Assumptions} The models demonstrate concerning biases in attributing emotional expressions (e.g., frustration, anger) to cultural norms of specific demographic groups. The expert emphasized that such reactions are universal human responses to situations like unresolved maintenance issues, not characteristics of any particular group.

\paragraph{Misinterpretation of Speech Patterns} The models incorrectly classify common features of verbal communication (e.g., short sentences, informal tone) as dialect-specific markers. However, these are typical characteristics of spoken language across all demographics.

\paragraph{Unfounded Assumptions About Language Sophistication} The models exhibit bias in equating informal language with a lack of sophistication, particularly problematic when associating this with specific demographic groups. As referenced by the expert, this misconception has been thoroughly addressed in seminal sociolinguistic works \citep{labov1969study}.

\end{document}